\documentclass[letterpaper,10pt,conference]{ieeeconf}

\IEEEoverridecommandlockouts
\usepackage{cite}
\usepackage{amsmath,amssymb}
\usepackage{graphicx}
\usepackage{booktabs}
\usepackage{array}
\usepackage{tabularx}
\usepackage{xspace}
\usepackage{microtype}
\usepackage{flushend}
\usepackage{xcolor}
\usepackage[hidelinks]{hyperref}

\newcommand{\sys}{\textsc{EmoPose}\xspace}
\newcommand{\revisioncolor}{black}
\newcommand{\rev}[1]{{\color{\revisioncolor}#1}}
\newcolumntype{Y}{>{\centering\arraybackslash}X}

\title{\sys: Vision-Language Model Guided Emotion-Aware\\
Gesture Generation for Humanoid Robots}

\author{Daojie Peng, Bingtao Wang, Fulong Ma, Wenjun Yue, Liang Zhang, Jun Ma$^\dagger$
\thanks{$\dagger$ Corresponding author: \texttt{jun.ma@ust.hk}}
\thanks{Daojie Peng, Fulong Ma and Jun Ma are with The Hong Kong University of Science and Technology (Guangzhou) (e-mail: \{fmaaf, dpeng108\}@connect.hkust-gz.edu.cn, jun.ma@ust.hk.)}
\thanks{Bingtao Wang and Liang Zhang are with The Shandong University, wangbt@mail.sdu.edu.cn, 201299800013@sdu.edu.cn 
}
\thanks{Wenjun Yue, is with the RoboScience. 
}
}

\begin{document}
\maketitle

\begin{abstract}
Socially competent humanoid robots must communicate affect and intent through gesture as well as speech, yet open-ended interaction must become motion that is both expressive and executable on a specific body. This demands \rev{semantic flexibility} for contextual social intent while preserving deterministic, embodiment-aware robot control. We present \sys, a vision-language model (VLM)-guided framework that bridges this gap through an \rev{executable semantic interface}. Given language, dialogue history, and optional visual context, the VLM selects an \rev{ordered gesture plan} containing a communicative class, library variant, intensity, and speech anchor. A scalable robot-owned motion library defines the available expressive vocabulary and the source of 14-DoF joint targets. Pose Studio supports \rev{automatic trajectory generation}, MuJoCo preview, and \rev{automatic synchronization} of new library entries with the VLM guide; deterministic robot-side modules validate plans, construct trajectories, schedule gestures, and manage queueing and interruption. This division lets the \rev{interaction repertoire grow} for new social contexts without changing the control interface or delegating raw joint commands to the foundation model. On the EmoPose-Bench, structured GPT-5.5 planning reaches $98.25\pm0.52\%$ on the Easy tier and $76.50\pm0.54\%$ overall, exceeding same-model direct-label prompting. 
Further tests validate dialogue-context use and ordered multi-action
composition.
The system completes the nominal MuJoCo suite
and realizes all 29 authored variants on the physical Unitree G1.
A four-stop laboratory tour demonstrates expressive narration
with interruption, camera-grounded dialogue, and navigation.
\end{abstract}

\section{Introduction}
For humanoid robots that guide, explain, or converse with people, speech is only part of the response. Gestures convey deixis, emphasis, stance, and affect that words may leave implicit~\cite{mcneill1992}, and robot body language shapes how people interpret and respond to an interaction~\cite{breazeal2003,saunderson2019}. A socially capable humanoid must therefore infer communicative intent, select a congruent gesture, coordinate it with speech, and realize it within the constraints of its own embodiment.

The \rev{core challenge} is a mismatch between two representations. Natural language is open-ended, contextual, and often indirect: agreement need not contain ``yes,'' reassurance may resemble welcome at the lexical level, and one reply may require several communicative acts in a specific order. Robot motion is finite and embodiment-specific, with explicit joint ranges, transition semantics, and interruption requirements. Fixed keyword mappings yield predictable actions but break under paraphrase and pragmatic implication. Learned co-speech and text-to-motion systems broaden motion diversity~\cite{yoon2019cospeech,kucherenko2020gesticulator,motiongpt2023}, while language-model planners increasingly compose expressive skills and humanoid gesture sequences~\cite{mahadevan2024genem,huang2024emotion}. However, neither a human-skeletal motion sequence nor free-form model output directly specifies a hardware-compliant \rev{robot command} for a particular humanoid. Physical deployment needs semantic flexibility while retaining \rev{robot-side joint control}.

\begin{figure}[t]
  \centering
  \includegraphics[width=0.92\linewidth]{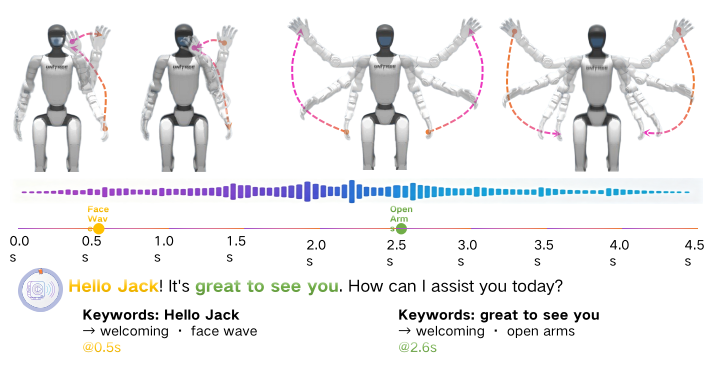}
  \caption{Two-gesture execution generated by \sys. The upper row overlays the library trajectories for \texttt{face\_wave} and \texttt{open\_arms}; the lower timeline shows their planned lexical releases relative to synthesized audio. The 0.5-s and 2.6-s release times are computed by Eq.~\eqref{eq:schedule}.}
  \label{fig:demo}
\end{figure}

We present \sys, a VLM-guided, emotion-aware gesture system built around a \rev{library-grounded semantic interface}. Given an utterance, dialogue history, and optional visual context, the VLM produces a verbal reply and an ordered sequence of records specifying gesture class, named variant, intensity, and lexical speech anchor. The \rev{active motion library} has two roles: it generates the action vocabulary exposed to the model and supplies every \rev{numerical joint target} at execution time. Deterministic robot-side modules validate and resolve each record, modulate library keyframes, construct trajectories, align gestures with synthesized speech, and manage blending, queuing, holding, and interruption. Thus, the VLM selects what the robot communicates, while the robot stack retains ownership of every commanded joint value. Figure~\ref{fig:demo} illustrates two library-grounded gestures released at different phrases within one spoken reply.

\sys provides a \rev{reusable semantic-to-motion interface} instead of a one-off language-to-motion mapping. Its live library couples the VLM vocabulary to executable robot motions, \rev{automatically synchronizing new trajectories} to support contextual dialogue, multi-gesture speech alignment, interruption, and navigation under deterministic control.

This paper makes three contributions:
\begin{itemize}
  \item \textbf{Executable semantic interface:} A live robot‑resident motion library that defines the valid gesture vocabulary for the VLM and serves as the source of joint‑space targets. This design enforces traceability for every motion command while enabling context‑aware, ordered gesture selection driven by high‑level language reasoning;
  \item \textbf{\rev{Extensible semantic-to-execution} system:} We implement a complete pipeline integrating Pose‑Studio‑based motion authoring, five‑level gesture intensity modulation, lexical‑anchor speech‑gesture alignment, deterministic trajectory generation, runtime gesture interruption handling, and combined gesture‑navigation interaction; 
  \item \textbf{Semantic and embodied evaluation:} We introduce \textit{EmoPose‑Bench}, a balanced utterance benchmark. Constructed with human‑assisted annotation, the benchmark covers three difficulty tiers (Easy, Medium, Hard) to test direct cue matching, paraphrase robustness, and pragmatic reasoning. We conduct controlled ablations for model variants, dialogue context, and multi‑action composition, with end‑to‑end validation over MuJoCo simulations, full‑library hardware tests on Unitree G1, and a four‑stop integrated laboratory tour demo.
\end{itemize}

Together, these contributions establish a reusable system pattern in which a foundation model enriches embodied
communication.

\section{Related Work}
\subsection{Expressive and Co-Speech Motion}
Classical planners align nonverbal acts with linguistic structure~\cite{cassell2001beat}, and communicative gestures have been authored and evaluated on social robots~\cite{salem2012}. Learned co-speech systems generate humanoid motion from acoustic and semantic cues~\cite{yoon2019cospeech,kucherenko2020gesticulator}, while BEAT and GENEA provide data and evaluation protocols~\cite{beat2022,genea2022}; text-to-motion models learn broader human-motion representations~\cite{motiongpt2023}. These approaches expand motion quality and diversity, but physical deployment still requires robot-specific joint limits, command traceability, transitions, and execution.

Foundation models increasingly select expressive behavior: GenEM composes parameterized skills~\cite{mahadevan2024genem}, while EMOTION generates gesture sequences~\cite{huang2024emotion}. These methods differ in input modality, embodiment, and evaluation target. \sys is complementary: one active library supplies both legal action names and numerical joint targets.

\subsection{Language Models as Robot Interfaces}
Language models can plan admissible action sequences, ground choices in skill affordances, and synthesize robot programs~\cite{huang2022zeroshot,saycan2022,liang2023code}. \rev{PaLM-E and RT-2} ground language and vision in sensorimotor representations~\cite{palme2023,rt2_2023}. 
Nevertheless, these methods focus largely on manipulation skills and often delegate raw motion generation to foundation models, which raises traceability and embodiment‑compliance challenges for expressive social‑gesture scenarios.
\sys addresses this gap by separating high‑level semantic planning from low‑level motion execution via a robot‑owned gesture library.

\section{Executable Gesture System}
\subsection{System Boundary and Semantic Record}
\begin{figure*}[t]
  \centering
  \includegraphics[width=0.95\textwidth]{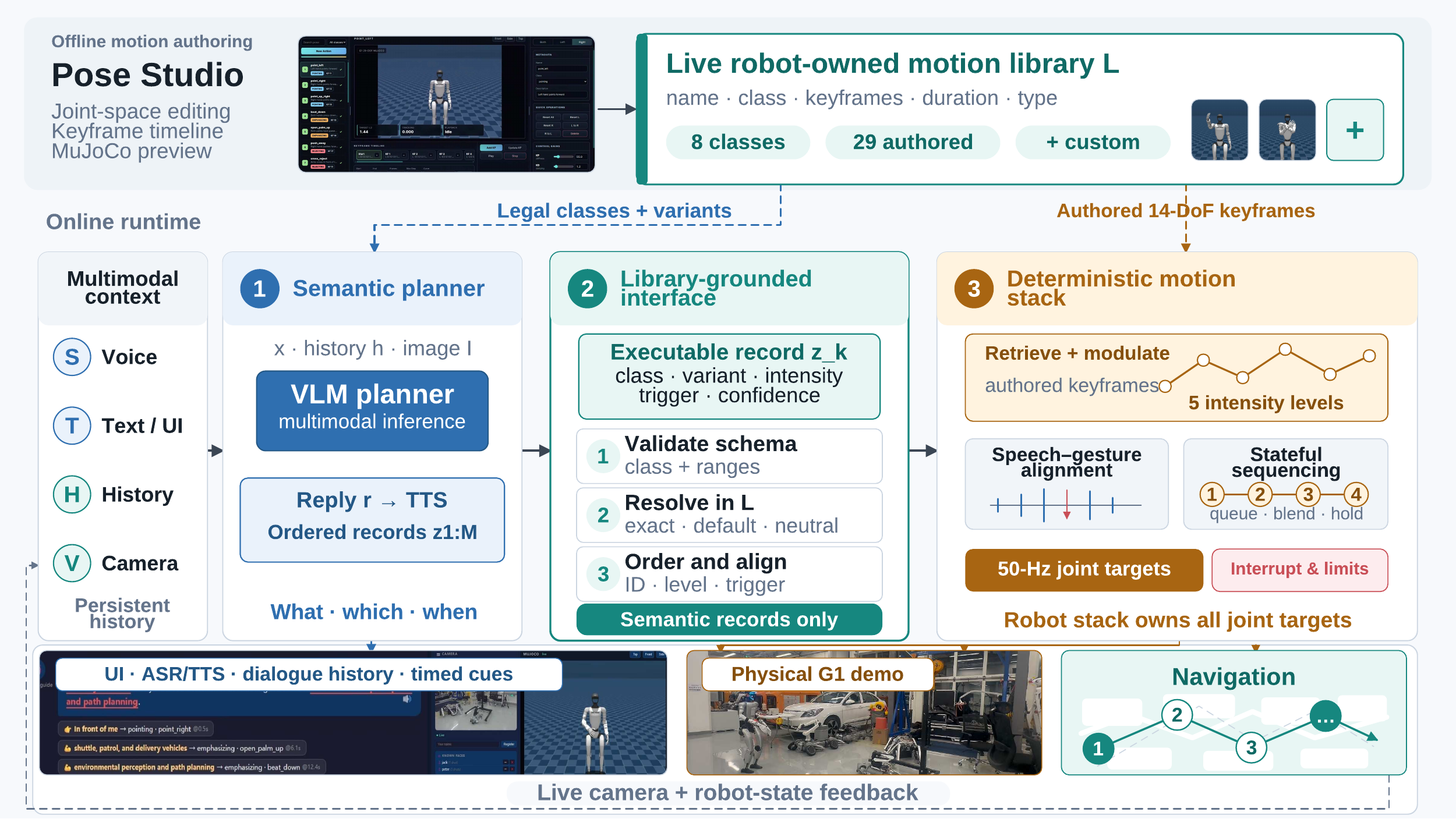}
  \caption{Architecture and \rev{semantic-to-motion control} boundary of \sys. Pose Studio builds the extensible robot-owned library $\mathcal L$, currently with \rev{29 motion variants and support for user-defined additions}. The library provides legal class-variant metadata to the VLM and 14-DoF keyframes to deterministic robot-side validation, resolution, modulation, speech alignment, sequencing, and safety. The runtime accepts ASR-transcribed voice, text/UI, dialogue history, and camera context, and produces TTS replies, MuJoCo and physical-G1 gestures, and navigation. 
  }
  \label{fig:architecture}
\end{figure*}

Figure~\ref{fig:architecture} presents \sys as a multimodal interaction-to-control architecture with an explicit \rev{semantic-numerical boundary}. At design time, Pose Studio \rev{automatically generates} and smooths trajectories, provides keyframe timing and \rev{MuJoCo preview}~\cite{todorov2012mujoco}, and registers named, class-tagged motions in the active robot-owned library. At run time, ASR-transcribed voice or typed text, dialogue history, and optional camera context enter the VLM semantic planner, while the same library regenerates its current \rev{class-variant guide}. The planner produces a verbal reply and ordered semantic records; TTS renders the reply for spoken interaction. Robot-side modules validate each record, resolve it against the live library, retrieve and modulate library keyframes, align releases with lexical triggers, synthesize and sequence trajectories, and dispatch gesture or navigation commands to MuJoCo and the physical G1. The Web UI maintains \rev{dialogue history}, timed gesture cues, \rev{live camera} and simulation feedback, physical execution, and guide-mode navigation.

Let the active motion library be
\begin{equation}
 \mathcal L=\{\ell_n=(n,c,Q,d,m)\},
 \label{eq:library}
\end{equation}
where $n$ is a unique variant name, $c\in\mathcal C$ is its semantic class, $Q=(q_0,\ldots,q_{K-1})$ contains 14-DoF keyframes, $d$ is duration, and $m$ is the motion type. \rev{Guided by the deictic and beat functions of co-speech gesture~\cite{mcneill1992} and the behavior-annotated SuSuInterActs corpus introduced in SentiAvatar~\cite{sentiavatar2026},} we instantiate $\mathcal C$ with eight communicative functions:
\begin{equation}
\begin{split}
\mathcal C=\{&\textit{pointing, emphasizing, rejecting, agreeing,}\\[-1mm]
              &\textit{doubting, welcoming, comforting, neutral}\}.
\end{split}
\end{equation}
\rev{The vocabulary covers reference and emphasis (pointing, emphasizing), conversational stance (agreeing, rejecting, doubting), social rapport (welcoming, comforting), and a neutral baseline.} \rev{It gives semantic planning a compact interface to an extensible robot-owned motion library: new variants can be added without changing the language-to-motion contract.}

For utterance $x$, optional dialogue history $h$, and a guide generated from $\mathcal L$, the VLM returns a reply $r$ and an ordered semantic record:
\begin{equation}
 z_k=(c_k,v_k,i_k,t_k,\gamma_k),
 \label{eq:record}
\end{equation}
with class $c_k$, variant name $v_k$, intensity $i_k\in\{1,\ldots,5\}$, lexical trigger $t_k$, and self-reported confidence $\gamma_k\in[0,1]$. The parser accepts bare or fenced JSON and clamps numeric fields. Unknown structured classes map to neutral; plain text yields a neutral record, whereas JSON decoding or value failures invoke a logged keyword fallback. 
The current resolver is
\begin{equation}
 \rho_{\mathcal L}(c,v)=
 \begin{cases}
  \ell_v, & v\in\operatorname{Name}(\mathcal L),\\
  \ell_c^{(0)}, & v\notin\operatorname{Name}(\mathcal L),\ \mathcal L_c\neq\varnothing,\\
  \ell_{\rm neutral}, & \mathcal L_c=\varnothing,
 \end{cases}
 \label{eq:resolver}
\end{equation}
where $\ell_c^{(0)}$ is the first configured primitive for class $c$. The resolver in Eq.~\eqref{eq:resolver} maps every semantic record to a concrete member of $\mathcal L$: exact-name lookup when available, a class-default fallback for a missing name, and neutral fallback for an empty class. Because the record retains both $c$ and $v$, class-variant agreement remains directly auditable.

This construction enforces a \rev{command-source} invariant in Eq.~\eqref{eq:authority}:
\begin{equation}
 q^{\rm cmd}_{0:T}=G\!\left(\rho_{\mathcal L}(c_k,v_k),\bar i_k,q^{\rm last}\right),
 \qquad \rho_{\mathcal L}(c_k,v_k)\in\mathcal L,
 \label{eq:authority}
\end{equation}
where $G$ is deterministic and $\bar i_k$ is the clamped intensity. Consequently, the VLM retains semantic choice over a finite motion vocabulary, while local robot software remains the exclusive source of every 14-DoF joint target. Table~\ref{tab:contract} summarizes the accepted fields and their deterministic local consequences.

\begin{table}[t]
\caption{Semantic-to-motion runtime contract.}
\label{tab:contract}
\centering
\scriptsize
\setlength{\tabcolsep}{2.8pt}
\begin{tabularx}{\columnwidth}{@{}lXX@{}}
\toprule
Field & Accepted input & Local consequence \\
\midrule
Class & eight labels & neutral on invalid label \\
Variant & any loaded name & global lookup; class fallback on miss \\
Intensity & integer 1-5 & clamp; Eq.~\eqref{eq:intensity} \\
Trigger & phrase or missing & schedule or interpolate \\
Confidence & scalar 0-1 & clamp and log only \\
Joint values & absent & always sourced from $\mathcal L$ \\
\bottomrule
\end{tabularx}
\end{table}

\subsection{Extensible Motion Library and Intensity}
The current deployment contains 29 motion sequences plus two built-in neutral poses, but the library is not capped at 31 entries. Pose Studio \rev{automatically generates} and smooths trajectories for new class-tagged entries, provides keyframe timing and MuJoCo preview, and \rev{synchronizes their metadata and keyframes} with the resolver and VLM guide. New behaviors thereby become available to semantic planning and execution \rev{without a separate language-to-action mapping}; platform-specific deployment supplies only a compatible library and limits. Motion counts in pointing, emphasizing, rejecting, agreeing, doubting, welcoming, comforting, and neutral order are $4,4,4,3,3,4,5,2$. Every library sequence contains $K=3$-14 keyframes over the seven joints of each arm. If a duration is not specified explicitly, it is computed as
\begin{equation}
 d(Q)=\operatorname{clip}\!\left(0.3(K-1),1,5\right)~\mathrm{s};
 \label{eq:duration}
\end{equation}
An explicit duration overrides Eq.~\eqref{eq:duration}. The same configuration feeds Pose Studio, the prompt guide, simulation, and physical execution, so \rev{library extensions propagate consistently}.

For a library sequence, intensity changes only interior excursion about the first keyframe. Let $\mathcal B=\{0,K-1\}$, and let $q_j^-$ and $q_j^+$ denote the configured lower and upper hard limits for joint $j$. Define the joint-wise projection
$\Pi_j(y)=\operatorname{clip}(y,q_j^-+\delta,q_j^+-\delta)$; then
\begin{equation}
 \tilde q_{k,j}^{(i)}=
 \begin{cases}
 q_{k,j}, & i=3\ \vee\ k\in\mathcal B,\\
 \Pi_j\!\left(q_{0,j}+a_i(q_{k,j}-q_{0,j})\right), & \text{otherwise},
 \end{cases}
 \label{eq:intensity}
\end{equation}
where $a_i\in\{0.4,0.6,1.0,1.3,1.6\}$ for intensity levels $i=1,\ldots,5$, respectively, and $\delta=0.02$~rad. Level 3 reproduces the library sequence exactly; the remaining levels preserve both boundary keyframes while scaling the interior expressive excursion. Duration remains invariant, and all 29 sequences contain at least one interior keyframe available for modulation.

\subsection{Trajectory Construction and Stateful Execution}
The configured trajectory path divides each keyframe segment into $M=5$ smoothstep increments (four strict interior samples and the endpoint):
\begin{equation}
 w_{k,r}=\tilde q_{k-1}+s(r/M)(\tilde q_k-\tilde q_{k-1}),\quad
 s(u)=3u^2-2u^3,
 \label{eq:smoothstep}
\end{equation}
for $r=1,\ldots,M$. The waypoint sequence generated by Eq.~\eqref{eq:smoothstep} is then linearly resampled at 50~Hz. The first segment begins at the latest published pose, preserving continuity from the robot's actual command state. Before publication, joint positions are projected into configured hard limits and a self-collision heuristic can reject unsafe candidates.

\begin{figure}[t]
  \centering
  \includegraphics[width=\columnwidth]{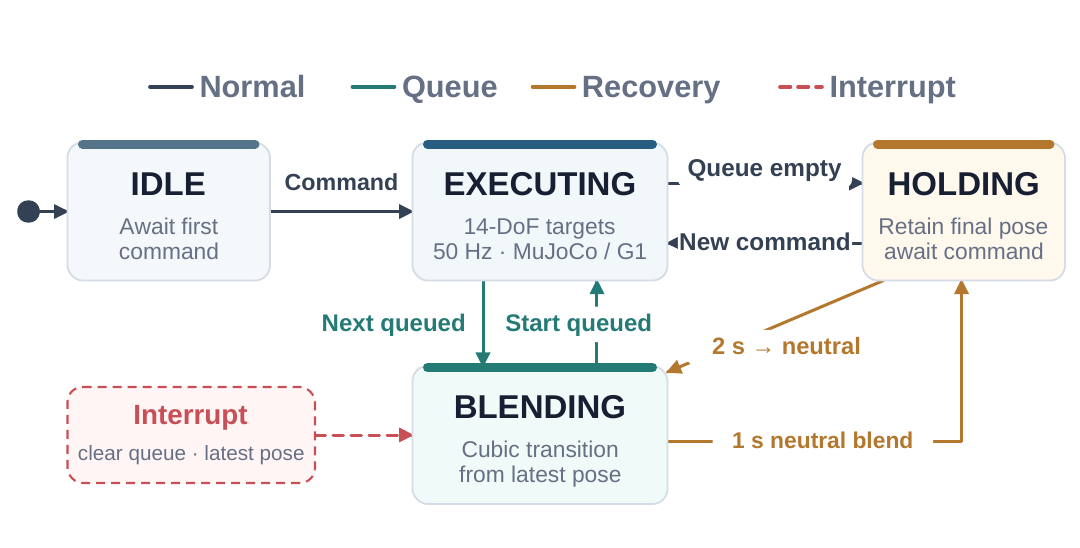}
  \caption{Robot-side queue-aware gesture sequencer. Dark arrows trace command dispatch and completion, teal paths handle queued successors, and amber paths show the 2-s hold and 1-s neutral recovery. The dashed coral path denotes interruption, which clears pending commands and restarts blending from the latest published pose. The 50-Hz rate applies to target publication in \textsc{executing}.}
  \label{fig:sequencer}
\end{figure}

The robot-side sequencer in Fig.~\ref{fig:sequencer} maintains one active trajectory and a FIFO queue. From \textsc{idle}, dequeuing a command constructs a trajectory that begins at the latest published configuration and enters \textsc{executing}; the initial transition is embedded in that trajectory. \textsc{executing} publishes 14-DoF targets at 50~Hz to MuJoCo or the physical G1. At completion, an empty queue leads to \textsc{holding}, whereas a queued successor passes through an explicit cubic \textsc{blending} transition of at most 0.3~s before execution. \textsc{holding} accepts a new command directly. After 2~s without one, a non-neutral endpoint is blended to neutral over 1~s and then held. When a new utterance interrupts execution, pending actions are cleared and a default 0.3~s blend starts from the latest published state toward the new target. The explicit blend curve is
\begin{equation}
 \begin{aligned}
 \beta(u)&=\begin{cases}
 4u^3,&u<\tfrac12,\\
 1-\tfrac12(-2u+2)^3,&u\geq\tfrac12,
 \end{cases}\\
 q_b(u)&=(1-\beta(u))q^{\rm last}+\beta(u)q^{\rm target}.
 \end{aligned}
 \label{eq:blend}
\end{equation}
Equation~\eqref{eq:blend} guarantees endpoint consistency: every transition starts from the latest commanded configuration and reaches its target without a positional discontinuity.

\subsection{Lexical-Anchor Speech Scheduling}
For synthesized audio of duration $T_a$, the first occurrence of trigger $t_k$ in reply $r$ schedules the center of that phrase at
\begin{equation}
 \hat\tau_k=
 \frac{\operatorname{pos}(t_k;r)+|t_k|/2}{\max(|r|,1)}T_a.
 \label{eq:schedule}
\end{equation}
When some anchors are absent, their release times are interpolated between neighboring matches. If no trigger matches and there are $M_g$ gestures, release $k$ is placed at $T_a(k+1/2)/M_g$; a single gesture is released immediately. This lightweight mechanism converts lexical anchors into reproducible release times using only reply text and synthesized-audio duration, enabling ordered multi-gesture replies without an external word-timing service.

\section{Experimental Evaluation}

\subsection{Semantic Selection, Composite Interface Ablation, and Error Structure}
\noindent\textbf{Protocol.}
EmoPose-Bench comprises 480 \rev{human-assisted} English utterances balanced over eight classes and three tiers (20 per class and tier). Its written construction rubric assigns \rev{direct lexical cues} to Easy, \rev{paraphrases without exact class names} to Medium, and \rev{pragmatic implication or competing cues} to Hard. Each sentence has one top-level target, although \sys may emit an ordered plan. We will \rev{release the utterances, labels, tier definitions, and evaluation code} with the project.

\begin{figure*}[t]
  \centering
  \includegraphics[width=0.9\textwidth]{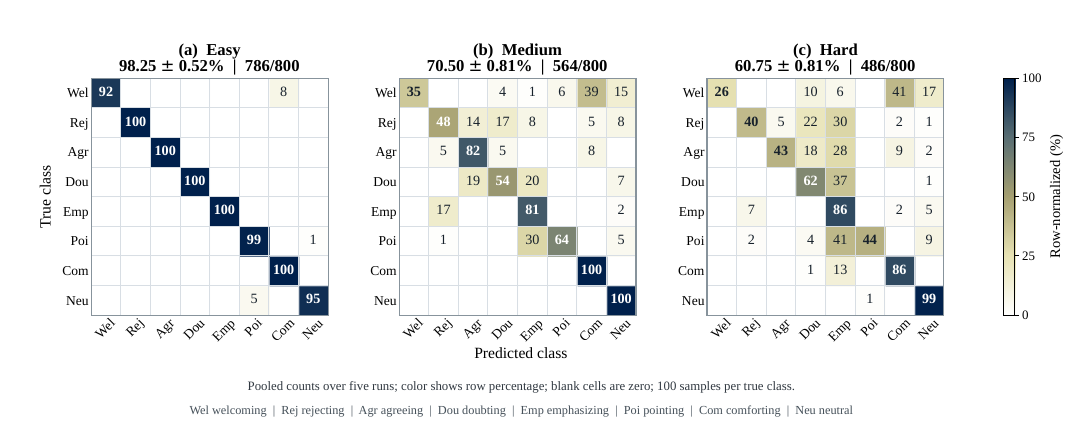}
  \caption{Pooled confusion matrices for five complete passes over 160 tier-specific utterances. Each panel aggregates 800 decisions; entries are pooled counts and color denotes row-normalized percentage. Titles report run-wise mean $\pm$ sample SD. Abbreviations: Wel welcoming, Rej rejecting, Agr agreeing, Dou doubting, Emp emphasizing, Poi pointing, Com comforting, Neu neutral.}
  \label{fig:confusion}
\end{figure*}

The full condition runs GPT-5.5 with class definitions, classification-priority rules, a live library guide, five in-context demonstrations, and the record in Eq.~\eqref{eq:record}; no demonstration utterance occurs in the benchmark. We issue one request per item at temperature 0.7 for five complete passes, taking the first planned class as top-1. 
Sentence-BERT uses \texttt{all-MiniLM-L6-v2} Easy-tier class centroids~\cite{reimers2019sentence}; keyword rules, TF-IDF, Jaccard, and chance provide lexical references. GPT-5.5 entries report run-wise mean and sample SD; Fig.~\ref{fig:confusion} pools the full condition's 2,400 predictions. 

\begin{table}[t]
\caption{Semantic benchmark and composite interface ablation: top-1 intent accuracy (\%). GPT-5.5 entries are mean $\pm$ sample SD over five passes; other entries are deterministic.}
\label{tab:semantic}
\centering
\scriptsize
\setlength{\tabcolsep}{2.3pt}
\begin{tabular}{@{}lrrrr@{}}
\toprule
Method & Easy & Medium & Hard & Overall \\
\midrule
\textbf{Ordered planning (ours)} & \textbf{98.25$\pm$.52} & \textbf{70.50$\pm$.81} & \textbf{60.75$\pm$.81} & \textbf{76.50$\pm$.54} \\
Direct-label GPT-5.5 & 95.00$\pm$.63 & 67.13$\pm$2.24 & 54.13$\pm$1.14 & 72.08$\pm$.74 \\
\textit{Full $-$ direct gain} & +3.25 & +3.37 & +6.62 & +4.42 \\
Sentence-BERT, Easy centroids & 97.50 & 58.75 & 52.50 & 69.58 \\
Keyword rules & 62.50 & 16.88 & 11.25 & 30.21 \\
TF-IDF descriptions & 26.88 & 22.50 & 23.13 & 24.17 \\
Jaccard descriptions & 21.25 & 17.50 & 18.13 & 18.96 \\
Uniform chance & 12.50 & 12.50 & 12.50 & 12.50 \\
\bottomrule
\end{tabular}
\end{table}



\noindent\textbf{Benchmark performance and composite‑interface ablation.}
Our full planner achieves an overall top‑1 accuracy of $76.50\pm0.54\%$ (Table~\ref{tab:semantic}). With the model, dataset, sampling strategy, and number of repeated runs held identical, the direct‑label ablation baseline obtains only $72.08\pm0.74\%$. The performance gap widens from 3.25 points on Easy to 6.62 points on the Hard tier, where linguistic cues are highly implicit. This result validates the benefit of our \rev{complete structured interface} while keeping the underlying model and benchmark unchanged.

Sentence‑BERT delivers strong performance ($97.50\%$) on the Easy tier it is fitted on, yet its accuracy drops sharply to $52.50\%$ for Hard‑level inputs. By comparison, simpler lexical baselines perform close to random chance. Our full planner outperforms all alternatives on the Medium, Hard, and overall metrics.

\noindent\textbf{Functional error structure.}
Figure~\ref{fig:confusion} localizes degradation to meaningful boundaries. Easy has only 14 errors among 800 decisions. The leading Medium confusions are welcoming$\rightarrow$comforting (39/100) and pointing$\rightarrow$emphasizing (30/100). On Hard these each occur 41 times, while doubting$\rightarrow$emphasizing occurs 37 times; comforting and neutral still retain 86\% and 99\% recall. The errors therefore concentrate at semantic boundaries between related communicative acts rather than at the motion interface itself. This result supports the paper's central design choice: preserve a compact, library-grounded motion contract while improving language-side distinctions with relational context and hierarchical intent representations.


\subsection{Dialogue‑History Ablation and Multi‑Action Planning}
\noindent\textbf{Protocol and metrics.}
We construct two balanced 320‑item test suites to evaluate the semantic record defined in Eq.~\eqref{eq:record}.
The context suite targets dialogue‑history ablation: it contains 40 final utterances for each gesture class, enabling paired evaluation with and without preceding dialogue context.
The sequence suite evaluates multi‑action composition; it consists of 80 two‑action, 160 three‑action, and 80 four‑action utterances, yielding 960 annotated action positions per experimental run.
All experiments adopt our structured GPT‑5.5 setup across five repeated passes.

Context‑aware classification accuracy is computed against ground‑truth annotated gesture classes for the context suite.
For the sequence suite, we report a suite of metrics: primary‑class accuracy, ordered exact match, aligned‑step accuracy, class‑multiset F1‑score, input‑anchor recall, and exact‑variant match.
Ordered exact match enforces perfect agreement with the full annotated class sequence, whereas exact‑variant match requires prediction of the precise library gesture name.
Input‑anchor recall quantifies how well the model recovers source trigger phrases from input utterances, serving as complementary assessment for the reply‑anchor scheduling logic in Eq.~\eqref{eq:schedule}.
Quantitative results from both suites are summarized in Table~\ref{tab:extensions}.

\begin{table}[t]
\caption{Dialogue-history ablation and ordered multi-action planning (\%; five-pass mean $\pm$ sample SD; length-specific exact reports means only).}
\label{tab:extensions}
\centering
\scriptsize
\setlength{\tabcolsep}{4.0pt}
\renewcommand{\arraystretch}{0.88}
\begin{tabularx}{\columnwidth}{@{}Xr@{}}
\toprule
Condition or metric & Score (\%) \\
\midrule
\multicolumn{2}{@{}l}{\textit{Dialogue-context ablation (same model)}} \\
Full planner (with history) & 96.69$\pm$.36 \\
Ablation (history removed) & 95.00$\pm$1.06 \\
Absolute gain from history & $+$1.69$\pm$.98 \\
No-overlap audit, full / ablated & 96.68$\pm$.36 / 94.98$\pm$1.06 \\
\addlinespace[1pt]
\multicolumn{2}{@{}l}{\textit{Ordered multi-action planning}} \\
Primary class & 97.75$\pm$.68 \\
Ordered exact & 91.50$\pm$.71 \\
Mean step class / class-multiset F1 & 95.49$\pm$.31 / 97.94$\pm$.28 \\
Input-anchor recall / annotated variant & 99.98$\pm$.03 / 44.88$\pm$.70 \\
Exact, length 2/3/4 & 96.75/94.00/81.25 \\
\bottomrule
\end{tabularx}
\end{table}

\begin{table}[t]
\caption{MuJoCo trajectory-execution results for the 50-Hz trajectory and sequencer path.}
\label{tab:execution}
\centering
\scriptsize
\setlength{\tabcolsep}{4.0pt}
\begin{tabular}{@{}lr@{}}
\toprule
Metric & Value \\
\midrule
Trials / motions / intensity & 160 / 8 defaults / level 3 \\
Completed trials & 160/160 \\
Stops / timeouts & 0 / 0 \\
Target-sim joint MAE & 0.04066~rad \\
Per-class MAE range & 0.02289-0.05420~rad \\
VLM inference mean / $P_{95}$ & 3.532 / 4.368~s \\
Trajectory construction mean / $P_{95}$ & 0.93 / 1.70~ms \\
Command submission mean / $P_{95}$ & 0.23 / 0.60~ms \\
\bottomrule
\end{tabular}
\end{table}

\begin{figure*}[t]
  \centering
  \includegraphics[width=0.92\textwidth]{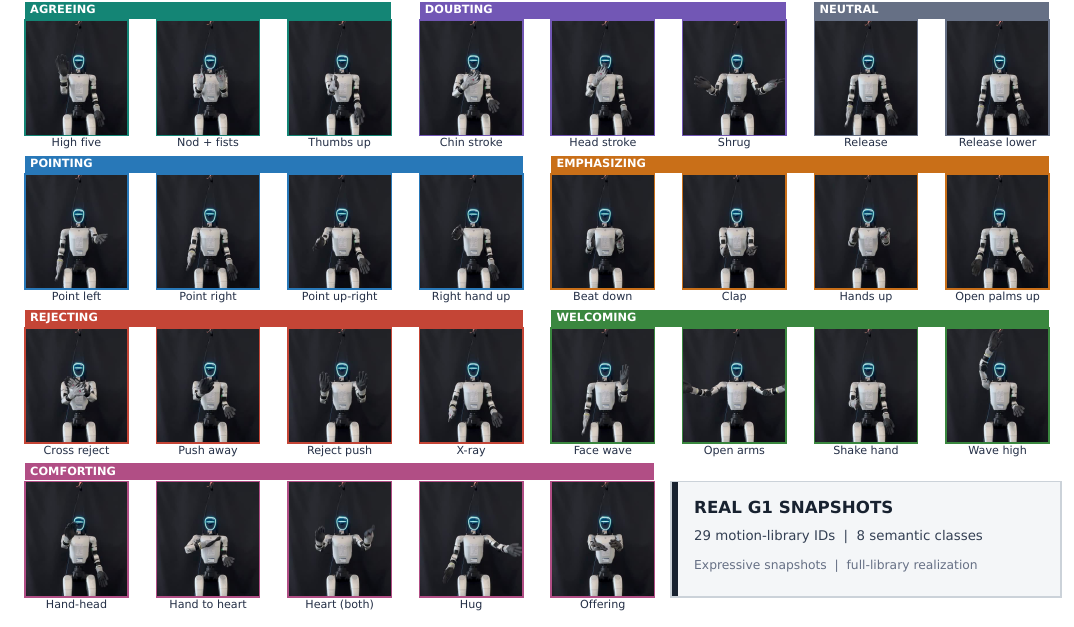}
  \caption{Qualitative full-library coverage on the physical G1. The 29 library motions span all eight communicative classes, with one representative frame from each recorded variant execution under the same stationary, overhead-supported protocol.}
  \label{fig:g1-library}
\end{figure*}

\begin{figure*}[t]
  \centering
  \includegraphics[width=0.9\textwidth]{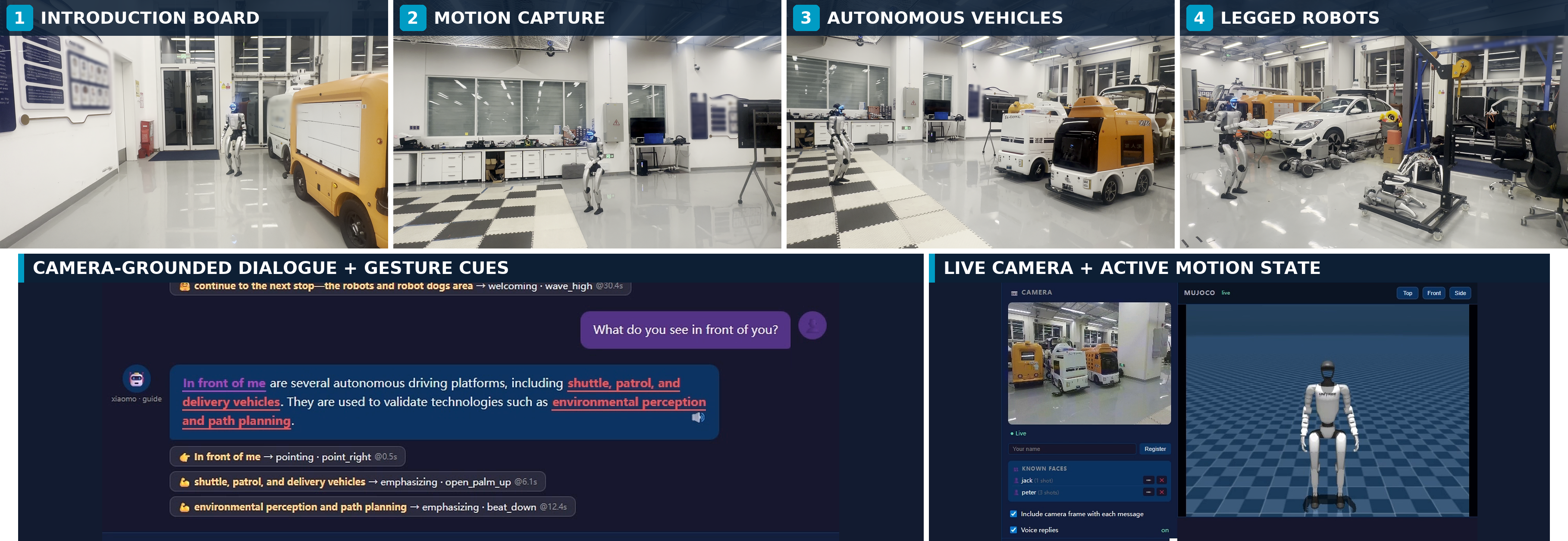}
  \caption{Integrated laboratory-tour demonstration. Top: frames from the final physical run at the introduction board, motion-capture area, autonomous-vehicle equipment area, and legged-robot equipment area. Enlarged views from the English-language interface recording show an interrupting camera-grounded question and its gesture-annotated response (bottom left), together with the live camera and active motion state (bottom right). 
  }
  \label{fig:lab-tour}
\end{figure*}


\noindent\textbf{Dialogue‑history ablation.}
This paired ablation isolates the contribution of dialogue memory: the model, structured semantic contract, library guidance, sampling configuration, and target utterances are held fixed, while only preceding conversation turns are discarded. Our full planner achieves $96.69\pm0.36\%$ accuracy with dialogue history, compared to $95.00\pm1.06\%$ without history (Table~\ref{tab:extensions}), corresponding to a $1.69\pm0.98$‑point performance gain. History boosts performance in four out of five repeated runs and yields a tie in the remaining run. Excluding the single sample overlapping with in‑context demonstrations retains this improvement ($96.68\pm0.36\%$ versus $94.98\pm1.06\%$). This outcome validates exposing dialogue state to semantic planning, which resolves referential and elliptical utterances without modifying the deterministic motion execution layer.

\noindent\textbf{Ordered‑plan results.}
For multi‑action sequence planning, we obtain $97.75\%$ primary‑class accuracy and $91.50\%$ ordered‑exact match accuracy. Ordered‑exact performance declines from $96.75\%$ for two‑action sequences down to $81.25\%$ for four‑action cases. Across 1\,600 predicted plans, 65 instances are over‑segmented and none are under‑segmented; specifically, 56 predictions add an extra fifth action for four‑action ground‑truth targets. This asymmetry indicates longer prompts tend to trigger plan elaboration rather than action omission, which motivates enforcing explicit plan‑length constraints. All 4\,800 reference‑aligned outputs select semantically consistent library gesture names, and annotated input‑anchor recall hits $99.98\%$. The strict variant‑match score stands at $44.88\%$, measuring agreement against one predefined reference gesture name.

\subsection{Simulation and Physical-G1 Evaluation}
\noindent\textbf{MuJoCo protocol.}
We conduct 160 class-balanced 50-Hz trials (20 per class), using one representative level-3 motion per class. Completion requires full trajectory termination without a stop request or timeout. Commanded and simulated streams are resampled to $N=100$ points over $D=160$ trials and $J=14$ joints:
\begin{equation}
 E_{\rm sim}=\frac{1}{DNJ}\sum_{d=1}^{D}\sum_{n=1}^{N}\sum_{j=1}^{J}
 \left|q^{\rm sim}_{dnj}-q^{\rm target}_{dnj}\right|.
 \label{eq:simmae}
\end{equation}
We separately record VLM inference, local construction, and nonblocking submission latency.

\noindent\textbf{Execution accuracy and latency.}
All 160 trials complete without a stop request or timeout (Table~\ref{tab:execution}). Equation~\eqref{eq:simmae} gives 0.04066~rad ($2.33^\circ$) MAE, ranging by class from 0.02289 to 0.05420~rad. Hosted inference dominates the initial response delay (3.532~s mean and 4.368~s $P_{95}$), whereas trajectory construction averages 0.93~ms and submission 0.23~ms. The semantic record isolates that hosted-model delay from the local motion path, allowing a \rev{cached or local planner} to replace the hosted service without changing trajectory generation or execution.

\noindent\textbf{Full-library physical coverage.}
Recorded Unitree G1 footage contains one execution of every library variant under a consistent stationary, overhead-supported protocol. Figure~\ref{fig:g1-library} presents representative frames for all 29 motions and eight classes, including unilateral, bilateral, crossed-arm, and elevated-arm expressions. One specification drives automatic trajectory generation, language guidance, resolution, MuJoCo, and physical publication; \rev{new entries expand the vocabulary} without a separate language-to-action mapping.

\noindent\textbf{Integrated laboratory-tour demonstration.}
The same stack guides visitors along the four fixed stops in Fig.~\ref{fig:lab-tour}. G1 onboard odometry drives the route, while \sys schedules narration gestures. A visitor can interrupt for dialogue or a scene description; a VLM interprets the live camera frame, and \sys answers through speech and library-grounded gestures. The demonstration tests the composition of navigation, visual grounding, dialogue, narration, and expressive motion with \rev{robot-side base and arm control}.

\section{Conclusion}
\sys establishes an executable semantic layer between contextual language and humanoid motion: the VLM selects communicative meaning and ordering, while a live robot-owned library and deterministic modules retain control of resolution, speech alignment, trajectory construction, interruption, and execution. The full planner reaches $76.50\pm0.54\%$ on EmoPose-Bench, 4.42 points above the bundled same-model direct-label condition; dialogue history adds $1.69\pm0.98$ points on contextual turns; and exact ordered multi-action planning reaches $91.50\pm0.71\%$. All 160 MuJoCo trials complete with 0.04066~rad target-to-simulated-state MAE, physical footage covers all 29 library motions, and the laboratory tour integrates \rev{visual dialogue, gesture, and navigation}. Pose Studio \rev{synchronizes new library trajectories} with the semantic and execution stack.

The current benchmark is in English only, the physical repertoire is limited to 14-DoF upper-limb gestures, and the tour follows fixed destinations using onboard odometry. Future work will add independently annotated multilingual data and held-out prompt development, extend the authority architecture to coordinated whole-body behavior, and couple it with more stable localization, navigation, and perception-aware replanning.

\bibliographystyle{IEEEtran}
\bibliography{robio,icra_additions}

\end{document}